\documentclass{IEEEtran4PSCC}
\ifCLASSINFOpdf
   \usepackage[pdftex]{graphicx}
\else
   \usepackage[dvips]{graphicx}
\fi
\makeatletter
\let\old@ps@headings\ps@headings
\let\old@ps@IEEEtitlepagestyle\ps@IEEEtitlepagestyle
\def\psccfooter#1{%
    \def\ps@headings{%
        \old@ps@headings%
        \def\@oddfoot{\strut\hfill#1\hfill\strut}%
        \def\@evenfoot{\strut\hfill#1\hfill\strut}%
    }%
    \def\ps@IEEEtitlepagestyle{%
        \old@ps@IEEEtitlepagestyle%
        \def\@oddfoot{\strut\hfill#1\hfill\strut}%
        \def\@evenfoot{\strut\hfill#1\hfill\strut}%
    }%
    \ps@headings%
}
\makeatother

\usepackage[cmex10]{amsmath}
\usepackage{cite}
\usepackage{multirow} 
\usepackage{booktabs}

\usepackage{array}
\usepackage[linesnumbered,ruled,vlined]{algorithm2e}
\let\oldnl\nl
\newcommand{\nonl}{\renewcommand{\nl}{\let\nl\oldnl}}

\usepackage[caption=false,font=normalsize,labelfont=sf,textfont=sf]{subfig}

\usepackage{caption}
\usepackage{xcolor}

\usepackage{hyperref}  
\usepackage[nameinlink,capitalise]{cleveref}

\begin{document}

\title{Transferable Graph Learning for Transmission Congestion Management via Busbar Splitting}

\author{
\IEEEauthorblockN{Ali Rajaei, Peter Palensky, Jochen L. Cremer}
\IEEEauthorblockA{Department of Electrical Sustainable Energy \\
Delft University of Technology\\
Delft, The Netherlands\\
a.rajaei, p.palensky, j.l.cremer\{@tudelft.nl\}
}
}


\maketitle

\begin{abstract}

Network topology optimization (NTO) via busbar splitting can mitigate transmission grid congestion and reduce redispatch costs. However, solving this mixed-integer nonlinear problem for large-scale systems in near-real-time is currently intractable with existing solvers. Machine learning (ML) approaches have emerged as a promising alternative, but they have limited generalization to unseen topologies, varying operating conditions, and different systems, which limits their practical applicability. This paper formulates NTO for congestion management considering linearized AC power flow, and proposes a graph neural network (GNN)-accelerated approach. We develop a heterogeneous edge-aware message passing GNN to predict effective nodes for busbar splitting actions as candidate NTO solutions. The proposed GNN captures local flow patterns, improves generalization to unseen topology changes, and enhances transferability across systems. Case studies show up to $10^4\times$ speed-up, delivering AC-feasible solutions within one minute and a 2.3\% optimality gap on the GOC 2000-bus system. These results demonstrate a significant step toward near-real-time NTO for large-scale systems with topology and cross-system generalization.


\end{abstract}



\section*{Nomenclature}

\begin{IEEEdescription}[\IEEEusemathlabelsep\IEEEsetlabelwidth{$P_abc,Q_abd$}]
\item[Indices and Sets]
\item[$i \in \mathcal{V} $] Index for substations (nodes).
\item[$b \in \mathcal{B}$] Index for busbars $\mathcal{B}=\{b1,b2\}$.
\item[$l_{ij} \in \mathcal{L} $] Index for transmission lines (from $i$ to $j$).
\item[$g \in \mathcal{G} $] Index for generators.
\item[$d \in \mathcal{D} $] Index for load demands.  
\item[$\mathcal{L}_i,\mathcal{G}_i,\mathcal{D}_i$] Elements $e \in E_i$ connected to substation $i$.
\item[$\mathcal{L}^{cong}$] Set of congested transmission lines.
\item[$\mathcal{V}^p$] Set of substations selected by the proximity filter.
\item[$\mathcal{V}^{cnd}$] Set of candidate substations selected by the GNN for potential busbar splitting.


\item[Parameters]
\item[$\hat{P}_d,\hat{Q}_d$] Active and reactive power demand.
\item[$\hat{P}_g,\hat{Q}_g$] Active and reactive power generation.
\item[$\underline{V}_i,\overline{V}_i$] Minimum and maximum squared voltage magnitude.
\item[$\overline{Z}^s$] Maximum number of busbar splitting. 
\item[$b_{l},g_{l}$] Susceptance and conductance of line $l$. 
\item[$\overline{S}_{l}$] Maximum apparent power flow of line $l$.
\item[$\phi_d$] Tangent of the power factor angle of loads. 
\item[$\gamma$] Congestion cost parameter. 
\item[$s^{th}$] Line loading penalization threshold. 
\item[$M^\theta,M^v$] Big-M parameters.

\item[Variables] 
\item[$P_{g,b},Q_{g,b}$] Auxiliary variables for power generation.
\item[$P_{d,b},Q_{d,b}$] Auxiliary variables for load demand.
\item[$P_{l,i},Q_{l,i}$] Active and reactive power flow of line $l$ from $i$ side. 
\item[$P_{l,i,b},Q_{l,i,b}$] Active and reactive power flow of line $l$ through busbar $b$ at substation $i$. 
\item[$P^L_{l,i},Q^L_{l,i}$] Active and reactive power loss of line $l$.
\item[$\theta_{i,b},V_{i,b}$] Voltage angle and squared voltage magnitude of substation $i$ busbar $b$.
\item[$\theta_{l,i},V_{l,i}$] Voltage angle and squared voltage magnitude at the end of line $l$ on the side of $i$. 
\item[$z_i$] Connection status of coupler in substation $i$ (0: open, 1: close). 
\item[$z_g, z_d$] Connection status of generator and demand to busbars ($0:b1, 1:b2$). 
\item[$z_{l,i}$] Connection status of line $l$ to substation $i$ busbars ($0:b1, 1:b2$).
\item[$\mathcal{O}^{CM}$] Congestion management objective.
\end{IEEEdescription}

\section{Introduction}
Congestion management (CM) in transmission networks is becoming increasingly challenging due to the growing integration of distributed energy resources and electrification across sectors. NTO, particularly substation reconfiguration via busbar splitting, is a promising approach to relieve congestion and reduce re-dispatch costs. However, solving this mixed-integer non-linear optimization problem for large-scale systems in near-real-time remains computationally challenging. As a result, operators often rely on expert knowledge or predefined manuals, which can lead to sub-optimal system performance \cite{marot2018expert}. Recent approaches have explored ML to identify effective topological actions in near-real-time, but most models lack scalability, remain system-specific and fail to generalize to new topologies, operating conditions, or unseen systems. This lack of generalization and transferability limits their practical use due to: 1) the need for retraining and costly data collection for each new system, 2) difficulty for operators to detect when models become outdated, and 3) performance drops under rapidly changing grid conditions. In this context, this paper proposes a GNN-based approach that exploits the locality of CM to scale to large-scale systems, generalize to topology changes, and transfer to unseen grids \cite{wu2025universal}.


Optimal substation reconfiguration (OSR) extends the optimal power flow (OPF) problem by introducing binary variables for breaker statuses, allowing changes to the network topology. Several works \cite{zhou2019bus, wang2023bus, van2023bus, sogol2023congestion} apply sensitivity analysis as a heuristic to identify candidate busbar splitting actions. These heuristic approaches assume a fixed substation configuration and focus solely on selecting splitting actions, but they do not provide an optimal substation configuration. \cite{heidarifar2015network,goldis2016shift} formulate DC-OSR as a MILP using node-breaker modeling and shift factors, respectively. \cite{morsy5506460configure} develops a heuristic approach to solve DC-OSR by focusing on local congested regions. Building on \cite{heidarifar2015network}, \cite{park2020optimal} propose a sparse tableau formulation of AC PF, while \cite{heidarifar2021optimal} develops a MISOCP formulation. However, these approaches either rely on heuristics, are limited to DC power flow, or become computationally intractable for large-scale systems.

Recent research has investigated supervised and reinforcement learning (RL) approaches for NTO \cite{van2025survey}. For instance, \cite{chauhan2023powrl} develops an RL agent based on proximal policy optimization, while \cite{matavalam2022curriculum, lehna2023compare} propose curriculum-based RL to improve learning performance. Other works such as \cite{dorfer2022power, meppelink2025hybrid} mitigate unforeseen action consequences by simulating future outcomes, but at the cost of a computationally intensive Monte-Carlo tree search. \cite{hassouna2025soft} proposes a GNN-based imitation learning approach to predict the probability of each action’s effectiveness (as a soft label) in reducing maximum overloading, and \cite{de2025generalizable} trains a heterogeneous GNN to imitate a rule-based expert agent for day-ahead operation and improve generalization to N-1 conditions. \cite{rabab2025flow} investigates a different application of NTO, aiming to maximize cross-zonal capability rather than to relieve congestion. In \cite{rabab2025flow}, two separate GNNs predict breaker statuses and DC flows for a given topology, followed by a repair optimization to enforce DC PF equations. Nevertheless, these approaches rely on reduced action spaces, cannot guarantee operational feasibility, and struggle with generalization and transferability. Here, we refer to \textbf{generalization} as the ability to perform well on unseen topologies or operating conditions within the same grid, and \textbf{transferability} as the ability of a trained model to be efficiently applied or adapted to different grids.

This paper formulates NTO for CM with linearized AC PF constraints (AC-NTO-CM) as a MIP. We propose an edge-aware GNN that quickly predicts effective splitting actions, which are used as candidate solutions for AC-NTO-CM and also prioritized as branching variables in the branch-and-bound MIP solver. The proposed GNN is fully localized to learn local PF patterns, while a proximity filter improves learning performance by focusing on the most relevant regions. Thus, the proposed approach addresses fundamental challenges in ML-based NTO: scalability, generalization to unseen topologies, and transferability to different systems. 

The main contributions are:
\begin{itemize}
    \item The MIP formulation of NTO for CM considering linearized AC equations.
    \item The proposed edge-aware GNN to predict busbar splitting candidates. The GNN’s expressiveness, the learning task design and the proximity filter, enable generalization to unseen N-k topologies and transferability across different power systems.
    \item Significant computational speed-ups while maintaining constraint feasibility and minor optimality gaps.
    \item A novel regression index that estimates the effectiveness of splitting actions, improving interpretability.
\end{itemize}

The paper is organized as follows. \cref{sec2:formulation} introduces the proposed NTO MILP formulation. \cref{sec3:learning} presents the GNN-based solution approach. \cref{sec4:results} presents the case studies, and \cref{sec5:conc} concludes the paper.

\section{Congestion Management via Busbar Splitting}
\label{sec2:formulation}

\begin{figure}[t]
    \centering \includegraphics[width=0.4\textwidth, keepaspectratio=true,trim={0cm 10.0cm 17.9cm 0cm},clip]{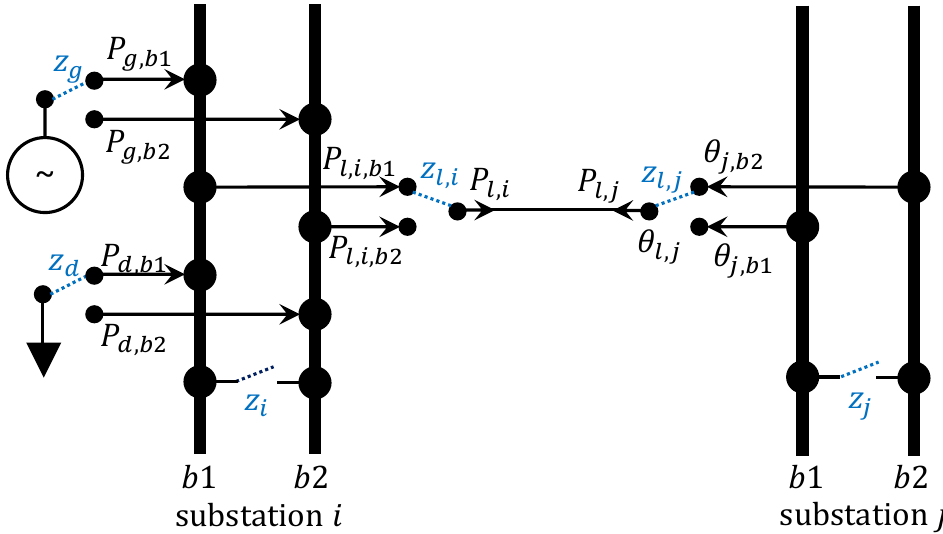}
    \caption{Schematic of the substation model for AC power flow.}
    \label{fig:substation}
\end{figure}

Substation reconfiguration via busbar splitting can effectively reduce transmission congestion by reconfiguring the grid to reroute flows, providing economic advantages over re-dispatch or load shedding measures. This paper focuses on two commonly used substation arrangements: the double-bus double-breaker and the breaker-and-a-half arrangements \cite{Book_sub_layout,heidarifar2015network,heidarifar2021optimal}. We extend the model in \cite{heidarifar2015network} to consider linear AC PF constraints. \cref{fig:substation} shows a substation $i$ connected to substation $j$ through a transmission line $l_{ij}$. Each  connected element $e \in E_i$ (lines, demands and generators) can be switched to either of the busbars $b1$ or $b2$ with the binary variable $z_e$, while $z_i$ denotes the busbar coupler status.

In this work, we consider the generation dispatch and load demand as input parameters. The constraints are:
\begin{subequations}
\label{eqCO:GD}
\begin{align}
&P_{g,b1} = (1-z_g)\hat{P}_g, \quad P_{g,b2} = z_g\hat{P}_g  \quad g \in \mathcal{G}  \label{eq:g0}\\
&Q_{g,b1} = (1-z_g)\hat{Q}_g, \quad Q_{g,b2} = z_g\hat{Q}_g  \quad g \in \mathcal{G}  \label{eq:g1} \\
&P_{d,b1} = (1-z_d)\hat{P}_d  ,\quad P_{d,b2}=z_d \hat{P}_d  \quad d \in \mathcal{D} \label{eq:d0}\\
&Q_{d,b}=\phi_d P_{d,b} \quad d \in \mathcal{D}\label{eq:d1} 
\end{align}
\end{subequations}

Each transmission line $l_{ij} \in \mathcal{L}_i$ can be connected to either busbars of substation $i$ through the binary variable $z_{l,i}$ as:
\begin{subequations}
\label{eqCO-lines}
\begin{align}
& -z_{l,i}M^\theta \leq \theta_{l,i} - \theta_{i,b1} \leq z_{l,i}M^\theta  \label{eq:lines-theta0}\\
& -z_{l,i}M^v \leq V_{l,i} - V_{i,b1} \leq z_{l,i}M^v \label{eq:lines-v0}  \\
& -(1-z_{l,i})M^\theta \leq \theta_{l,i} - \theta_{i,b2} \leq (1-z_{l,i})M^\theta  \label{eq:lines-theta1} \\
& -(1-z_{l,i})M^v \leq V_{l,i} - V_{i,b2} \leq (1-z_{l,i})M^v   \label{eq:lines-v1} 
\end{align}
\begin{align}
& -(1-z_{l,i})\overline{S}_{l} \leq P_{l,i,b1}  \leq (1-z_{l,i})\overline{S}_{l}   \label{eq:lines-P0} \\
& -(1-z_{l,i})\overline{S}_{l} \leq Q_{l,i,b1}  \leq (1-z_{l,i})\overline{S}_{l}  \label{eq:lines-Q0}  \\
& -z_{l,i}\overline{S}_{l} \leq P_{l,i,b2}  \leq z_{l,i}\overline{S}_{l} \qquad \qquad \qquad  \qquad\label{eq:lines-P1} \\
& -z_{l,i}\overline{S}_{l} \leq Q_{l,i,b2}  \leq z_{l,i}\overline{S}_{l}   \label{eq:lines-Q1} \\
& P_{l,i}=P_{l,i,b1}+P_{l,i,b2} \qquad \qquad \qquad  \qquad  \\
& Q_{l,i}=Q_{l,i,b1}+Q_{l,i,b2}  \label{eq:lines-Ql}
\end{align}
\end{subequations}
where $\theta_{i,b}$ is the voltage angle of the substation $i$ on each busbar $b$, and $\theta_{l,i}$ is the voltage angle of line $ij$ on the $i$ side, as \cref{fig:substation} shows. 

Constraints \eqref{eq:zmax}-\eqref{eq:z-sym2} consider operational assumptions of substations and improve the computational complexity:
\begin{subequations}
\label{eqCO-sw}
\begin{align}
&\sum_{i \in \mathcal{V}} (1-z_i) \leq \overline{Z}^s \label{eq:zmax}\\
&2(1-z_i) \leq \sum_{l \in \mathcal{L}_i}z_{l,i} \quad i \in \mathcal{V} 
\label{eq:z-sym0} \\
&2(1-z_i) \leq \sum_{l \in \mathcal{L}_i}(1-z_{l,i}) \quad i \in \mathcal{V} \label{eq:z-sym1} \\
&z_i + z_e \leq 1 \quad e \in E_i \label{eq:z-sym2} 
\end{align}
\end{subequations}
where \eqref{eq:zmax} limits the total number of busbar splitting actions, as set by the system operator. \eqref{eq:z-sym0}-\eqref{eq:z-sym1} ensure system security by enforcing that at least two lines are connected to each busbar if that substation splits, which prevents busbar isolation due to a line contingency \cite{heidarifar2015network, heidarifar2021optimal}. When the busbars are not split, the substation operates as a single node, and \eqref{eq:z-sym2} removes the symmetry in the formulation by enforcing all elements to connect to $b1$ in that case.

AC PF equations are crucial for accurately representing modern power systems, whereas DC formulations neglect reactive power and voltage constraints, which can lead to infeasible or sub-optimal solutions. However, using full AC equations results in a computationally expensive MINLP problem. In this paper, we use the linearized AC approximation \cite{alizadeh2022tractable} to achieve a balance of accuracy and efficiency in a MILP formulation: 
\begin{subequations}
\label{eqCO-PF}
\begin{align}
& \sum_{g \in \mathcal{G}_i} P_{g,b} - \sum_{d \in \mathcal{D}_i}P_{d,b}  = \sum_{l \in \mathcal{L}_i} P_{l,i,b}  \quad i \in \mathcal{V}, b \in \mathcal{B} \label{eq:PF-Pbalance} \\
&\sum_{g \in \mathcal{G}_i} Q_{g,b} - \sum_{d \in \mathcal{D}_i} Q_{d,b} = \sum_{l \in \mathcal{L}_i} Q_{l,i,b}  \quad i \in \mathcal{V}, b \in \mathcal{B} \label{eq:PF-Qbalance} \\
&P_{l,i}=g_{i}V_{l,i}+ g_{l}\frac{V_{l,i}-V_{l,j} }{2} - b_{l}(\theta_{l,i}-\theta_{l,j}) \nonumber \\ &+ P^L_{l,i} \quad l_{ij} \in \mathcal{L} \label{eq:Pij-AC} \\
&Q_{l,i}=-b_{i}V_{l,i}- b_{l}\frac{V_{l,i}-V_{l,j} }{2} - g_{l}(\theta_{l,i}-\theta_{l,j}) \nonumber \\ &+ Q^L_{l,i} \quad l_{ij} \in \mathcal{L} \label{eq:Qij-AC} 
\\
& P^2_{l,i}+Q^2_{l,i} \leq \overline{S}^2_{l} \quad l_{ij} \in \mathcal{L}  \label{eq:linelimit} \\
& \underline{V}_i  \leq V_{i,b} \leq \overline{V}_{i}  \quad i \in \mathcal{V},  b \in \mathcal{B} \label{eq:Vmax} 
\end{align}
\end{subequations}
where \eqref{eq:PF-Pbalance}-\eqref{eq:PF-Qbalance} present power balance for each substation's busbars. Active and reactive power flows are defined in \eqref{eq:Pij-AC}-\eqref{eq:Qij-AC}, with $P^L_{ij,c},Q^L_{ij,c}$ representing active and reactive power losses.  \eqref{eq:linelimit} models the thermal limit of lines. \eqref{eq:Vmax} represents voltage limits. We use a piecewise linearization of quadratic constraint \eqref{eq:linelimit}, and a linear approximation of loss terms $P^L_{l,i},Q^L_{l,i}$ around an initial operating point as in \cite{alizadeh2022tractable}.

The NTO for CM (denoted as AC-CM-NTO) is written as \eqref{eq:CO}, with the objective aiming to minimize grid congestion:
\begin{equation}
    \label{eq:CO}
         \begin{split}         
\min_{z_.} & \quad  \mathcal{O}^{CM}=\sum_{l \in \mathcal{L}} f( \left| \frac{S_l}{\overline{S}_{l}} \right|^{\gamma}) 
 \\  \text{s.t.}& \quad \text{\eqref{eqCO:GD}-\eqref{eqCO-PF} .} 
     \end{split}
\end{equation}
where $\gamma \geq 1$ is a parameter, and $f(.)$ is a convex function. The objective penalizes normalized line apparent flows (i.e., current magnitudes) \cite{sogol2023congestion}. In this work, we use $f(.)=\text{clip}(.,s^{th},\infty)$, to restrict penalization to lines whose loading exceeds a specified threshold $s^{th}$ (e.g., 80\%), whereas the parameter $\gamma\geq2$ shapes the penalty function by emphasizing highly loaded lines and down-weighting lightly loaded ones. We apply a piecewise linearization of the objective to obtain a MILP formulation.

\begin{figure}[t]
\centering
\includegraphics[width=0.5\textwidth]{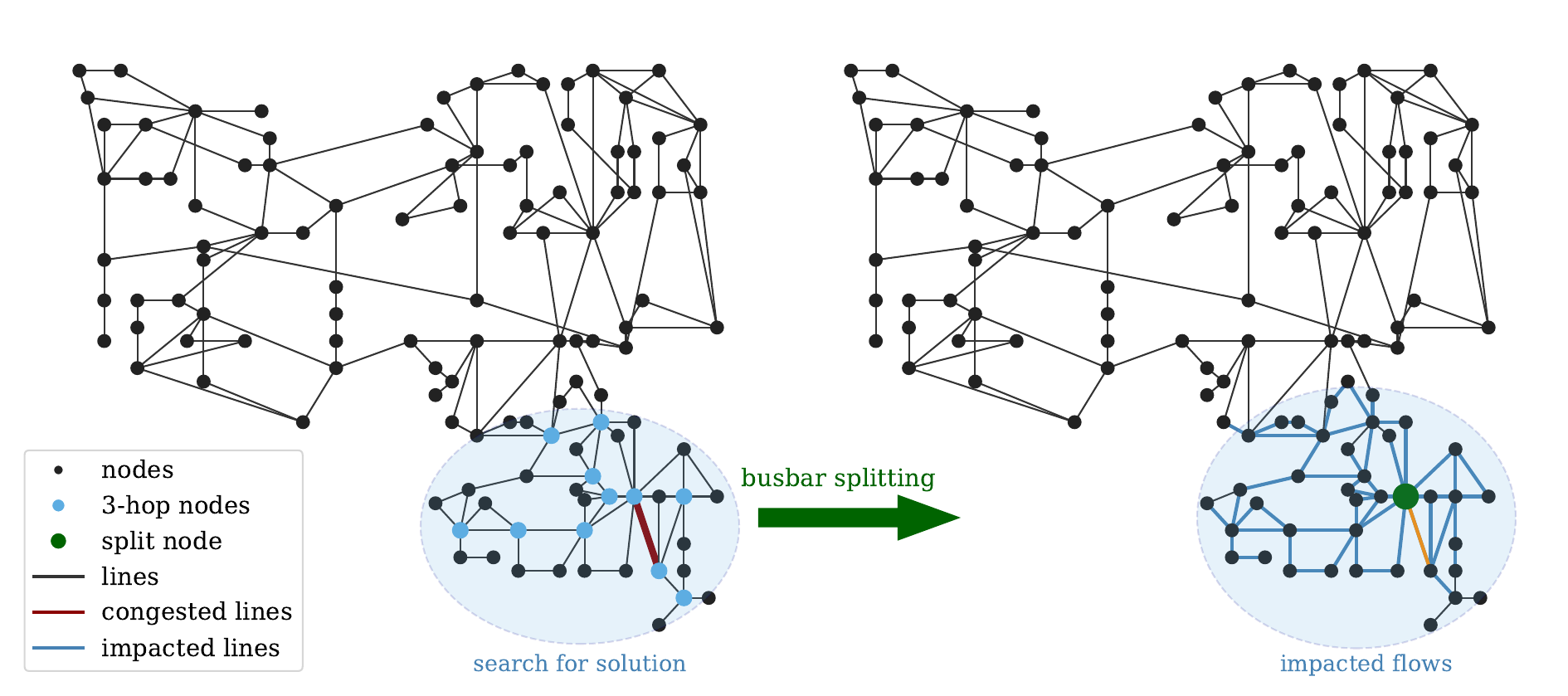}
\caption{An example of congestion in the IEEE 118-bus system. Left: potential busbar splitting actions within 3 hops. Right: line flows impacted by the splitting action.}
\label{fig:CM_network}
\end{figure}

\section{Transferable Graph Learning Approach}
\label{sec3:learning}

\begin{figure}[t]
    \centering \includegraphics[width=0.47\textwidth, keepaspectratio=true,trim={0cm 8.2cm 13.5cm 0cm},clip]{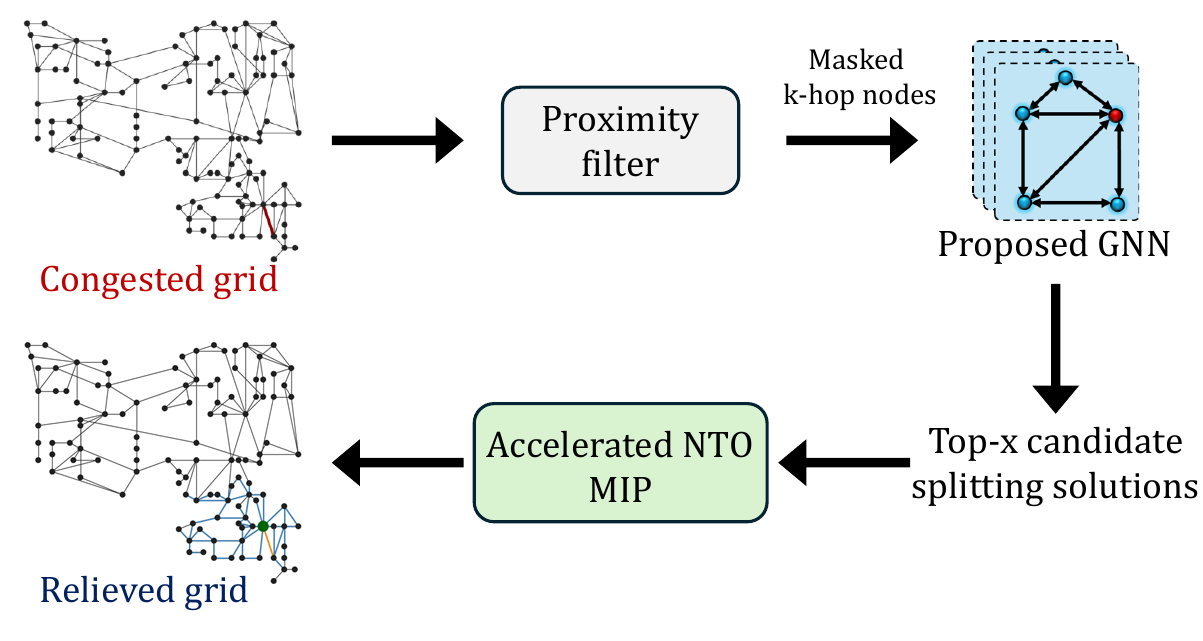}
    \caption{Workflow of the proposed GNN-accelerated NTO approach.}
    \label{fig:method_over}
\end{figure}

\subsection{GNN-accelerated NTO approach}

This section presents the proposed GNN-accelerated NTO (GNN-NTO) approach which significantly reduces the computational complexity of \eqref{eq:CO}, enabling near-real-time solutions while ensuring linear AC PF constraints. As shown in \cref{fig:CM_network} and previous studies \cite{meppelink2025hybrid, morsy5506460configure}, a busbar splitting action mostly impacts its local region, so effective actions are typically in close proximity to the congested line. This locality property of the NTO-CM problem makes GNNs a natural fit, as they use local information exchange and can improve scalability, generalization and transferability. 

\cref{fig:method_over} and \cref{alg:gnn-nto} present the workflow of the proposed approach, consisting of three stages: proximity filter, GNN-based candidate prediction, and accelerated NTO.

\subsubsection{Proximity Filter}
Given a congested grid state, the first step identifies substations likely to influence congestion. Congested lines $\mathcal{L}^{cong}$ are detected based on a threshold $s^{th}_l$ (Line~\ref{alg:cong}). A proximity filter then selects substations within $k$ hops (e.g., 5 hops) of the congested lines and with at least four connected lines, as defined in \eqref{eq:z-sym1}--\eqref{eq:z-sym2}, forming the proximity node set $\mathcal{V}^p$ (see \cref{fig:CM_network}). The filter does not remove input features; instead, it restricts the GNN output space to substations more likely to affect congestion, improving learning efficiency. During training, backpropagation is also applied only to these nodes.

\subsubsection{GNN-Based Candidate Prediction}
The grid state is encoded as a graph and processed by the proposed GNN, which predicts a splitting score $\hat{y}_i$ for each substation $i \in \mathcal{V}^p$ (Line~\ref{alg:gnn_forward}). These scores indicate how promising splitting a substation is for congestion mitigation. Substations are then ranked based on $\hat{y}_i$, and the top-$x$ (e.g., top-3) are selected as candidate nodes $\mathcal{V}^{cnd}$ (Line~\ref{alg:top-x}).

\subsubsection{Accelerated NTO}
The predicted candidate substations are incorporated into the NTO optimization. Substations outside the candidate set are fixed as unsplit nodes ($z_i = 1 \text{ } i \notin \mathcal{V}^p$), significantly reducing the number of binary variables (Line~\ref{alg:fix}). Moreover, candidate variables $z_i \text{ } i \in \mathcal{V}^p$ are prioritized as \textit{branching} variables in the branch-and-bound (B\&B) process of the MIP solver \cite{gurobi} (Line~\ref{alg:branch}). In other words, these variables are assigned higher branching priority compared to other binary variables (e.g., $z_{l,i}, z_g, z_d$), guiding the solver to explore promising topology decisions earlier in the search process. The reduced problem, denoted NTO$^*$, is then solved using a commercial solver (Line~\ref{alg:solve}).

By focusing the search on high-quality candidates predicted by the GNN, the proposed approach significantly accelerates the NTO-CM solution while maintaining high solution quality.

\begin{algorithm}[t]
\caption{GNN-NTO Approach}
\label{alg:gnn-nto}

\KwIn{$\hat{P}_{d},\hat{Q}_{d}$ $d\in \mathcal{D}$, $\hat{P}_{g},\hat{Q}_{g}$ $g\in \mathcal{G}$, $P_l,Q_l$ $l \in \mathcal{L}$}

\label{alg:cong} Detect congested lines 
$\mathcal{L}^{cong}=\{l \in \mathcal{L} \mid S_l \ge s^{th}\}$\;

\nonl \textbf{Proximity Filter} \\
\label{alg:prox_start}

\For{$l \in \mathcal{L}^{cong}$}{
    \For{$i \in \mathcal{V}$}{
        \If{$dist(l,i)\le k$ \textbf{and} $Deg(v) \ge 4$}{
            add $i$ to $\mathcal{V}^p$\;
        }
    }
}
\label{alg:prox_end}
\nonl \textbf{GNN-Based Candidate Prediction} \\ Forward pass of GNN 
$\hat{y}_i=\phi^{GNN}(x,e) \quad i \in \mathcal{V}^p$\label{alg:gnn_forward}\;

Select the top-$x$ substations $i$ from $\mathcal{V}^p$ ranked by $\hat{y}_i$, and add them to $\mathcal{V}^{cnd}$\label{alg:top-x}\;

\nonl \textbf{Accelerated NTO}\;

Fix non-candidate substations $z_i = 1 \quad  i \notin \mathcal{V}^{cnd}$ \;
\label{alg:fix}

Prioritize candidate variables $z_i \quad  i \in \mathcal{V}^{cnd}$ in B\&B\;
\label{alg:branch}

Solve reduced NTO$^*$ problem \eqref{eq:CO}\;
\label{alg:solve}
\Return optimal topology\;
\end{algorithm}

\begin{figure*}[t]
    \centering \includegraphics[width=0.8\textwidth, keepaspectratio=true,trim={0cm 12.3cm 0cm 0cm},clip]{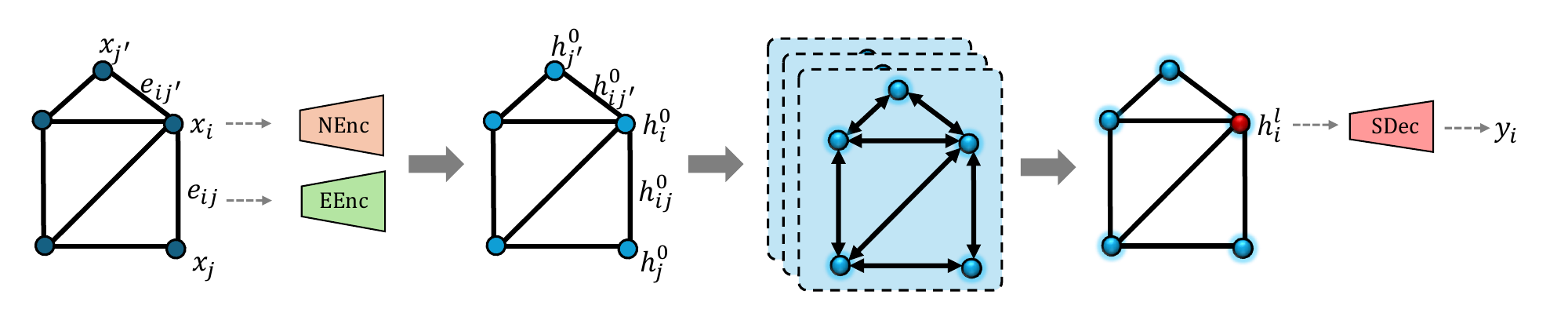}
    \caption{Schematic of the proposed encoder-GNN-decoder architecture.}
    \label{fig:architecture}
\end{figure*}

\subsection{Learning Task}
\label{sec:task}
This paper investigates two learning objectives: (1) \textit{whether} to split a node, modeled as classification (Clf), and (2) \textit{how much} congestion is reduced by splitting a node, modeled as regression (Reg). In both settings, the effect of splitting substation $i$ is evaluated independently, assuming that all other substations remain unsplit.

Congestion can be reduced by reconfiguring substation $i$ as:
\begin{equation}
\label{eq:sub_i}
    \begin{split}
        \min_{z_i,\{z_e\}_{e \in E_i}   }   \mathcal{O}^{CM}_i \\
        \text{s.t. } \eqref{eqCO:GD}-\eqref{eqCO-PF}.
    \end{split}
\end{equation}
where \eqref{eq:sub_i} solely solves for the binary variables of elements of substation $i$. Consider $\mathcal{O}^{CM}_0$ as the initial objective without any topological actions. Then, the congestion reduction by splitting substation $i$ is given by $\Delta\mathcal{O}^{CM}_i := \mathcal{O}^{CM}_0 - \mathcal{O}^{CM}_i$.

Let $y_i$ denote the GNN output label for substation $i$. In the Clf task, the goal is to predict \textit{whether} splitting substation $i$ reduces congestion. Therefore, $y_i^{Clf}=1$ if $\Delta\mathcal{O}^{CM}_i > \mathcal{O}^{th}$, and $y_i^{Clf} = 0$ otherwise, where $\mathcal{O}^{th} \ge 0$ is a pre-defined threshold.

In the Reg task, the goal is to predict \textit{how much} congestion reduction is achieved by splitting substation $i$, expressed as $y_i^{Reg} = \Phi( \Delta\mathcal{O}^{CM}_i )$, where $\Phi$ is a convex monotonic function. The Reg task provides a richer estimate of action effectiveness, allowing better interpretability and prioritization of candidate actions by operators/MIP solvers. However, it requires hyperparameter tuning of $\gamma$ in \eqref{eq:CO} and selecting $\Phi$. In this work, we use a clipping function $\Phi(.)=\text{clip}(.,\Delta\mathcal{O}^{low}_i, \Delta\mathcal{O}^{high}_i)$ to limit the influence of extreme values and stabilize the learning.

\subsection{Transferable Heterogeneous GNN Architecture}

We propose a transferable fully-local GNN approach to predict busbar splitting decisions on the substation level. \cref{fig:architecture} shows the proposed encoder-GNN-decoder architecture. Node and edge features are first processed through their respective encoders (each shared among all nodes or all edges) to obtain node and edge embeddings. The GNN layer then performs local message passing, followed by a splitting decoder that predicts the busbar splitting decision. Notably, the architecture relies solely on localized computations and is independent of the specific system or topology, i.e., the number of model parameters does not change with system size. 



\subsubsection{Edge-Aware Message Passing}
\label{sec:MPNN}

This section presents the homogeneous edge-aware message passing neural network (MPNN). The transmission grid is modeled as a graph $\mathcal{G} = \{\mathcal{V}, \mathcal{L}\}$, where each node $i \in \mathcal{V}$ corresponds to a substation, and each edge $(i, j) \in \mathcal{L}$ represents a transmission line $l_{ij}$. Each line is represented as a pair of bi-directional edges to enhance expressiveness and capture flow symmetry.

Consider the nodal feature as $x_i=[p_i,q_i,V_i,\text{Deg}_i,\text{Dis}_i]$, where $p_i,q_i$ are the net power injections (generation minus demand), $V_i$ is voltage magnitude, $\text{Deg}_i$ is the number of connected edges, and $\text{Dis}_i$ is the number of hops to the closest congestion. The edge feature is defined as $ e_{i,j}=[P_{l,i},Q_{l,i},S_{l,i},\overline{S}_{l},r_l,x_l ]$. Let $h^l_i$ and $h_{i,j}^l$ denote the node and edge embeddings at layer $l$. The proposed MPNN is:
\begin{subequations}
\label{eq:GNN}
    \begin{align}
        & h^0_i = \phi^{NEnc}(x_i)  &i \in \mathcal{V} \label{eqGNN-1}\\
        & h^0_{i,j} = \phi^{EEnc}(e_{i,j}) &ij \in \mathcal{L} \label{eqGNN-2}\\
        & \tilde{h}^l_{i,j} = \phi_\text{msg}^l( [ h^{l-1}_{i} || h^{l-1}_{j} || h^{l-1}_{i,j} ] )  &ij \in \mathcal{L} \label{eqGNN-3}\\
        & h^l_{i,j} = h^{l-1}_{i,j} + \tilde{h}^{l}_{i,j}  &ij \in \mathcal{L} \label{eqGNN-4} \\
        & m^l_i = \sum_{j \in N_i}( h^l_{i,j} )  &i \in \mathcal{V} \label{eqGNN-5}\\
        & h^l_i = h^{l-1}_i + \phi_\text{upd}^l( [ h^{l-1}_{i} || m^{l}_{i} ]   )   &i \in \mathcal{V} \label{eqGNN-6} \\
        & \hat{y}_i = \phi^{SDec}(h^L_i) &i \in \mathcal{V}^p \label{eqGNN-7}
    \end{align}
\end{subequations}
Here, $\phi(.)$ shows a two-layer MLP with ReLU activation, and $||.||$ denotes concatenation. \eqref{eqGNN-1}-\eqref{eqGNN-2} show the node and edge encoders, respectively. Edge messages are constructed in \eqref{eqGNN-3}, and the edge embeddings are updated with a residual connection in \eqref{eqGNN-4}. Messages are aggregated in \cref{eqGNN-5}, and node embeddings are updated in \eqref{eqGNN-6}, using a residual connection. After $L$ layers of message passing, the split decoder (SDec) in \eqref{eqGNN-7} outputs the final splitting decision. As mentioned earlier, we apply a proximity filter to improve learning performance: nodes $i \in \mathcal{V}^p$ are masked for prediction and used for backpropagation.


\subsubsection{Heterogeneous GNN}
\label{sec:HetGNN}
We extend the homogeneous MPNN in \eqref{eq:GNN} to a heterogeneous design that differentiates among node and edge types, providing a richer representation of the underlying power grid. To this end, we consider four node types: generators $g \in \mathcal{G}$, demands $d \in \mathcal{D}$, splitting nodes $i \in \mathcal{V}^p$, and non-splitting nodes $i \notin \mathcal{V}^{p}$. The splitting nodes are determined by the proximity filter for each grid sample. Moreover, we consider two edge types: transmission line $l_{ij} \in \mathcal{L}$, and pseudo connectors that link generators and demands to nodes. The pseudo connectors are featureless. We use separate $\phi_t(.)$ functions to encode and process the embeddings of each node type $t$.

\subsection{Training Data Generation}
\cref{alg:data} summarizes the training data generation process. First, random demands, planned $N\!-\!k$ line outages, and generation costs are sampled to emulate diverse operating conditions. The LP-AC-OPF is then solved to obtain the generation dispatch and initial power flows. To improve learning performance, only congested grid states are retained for training \cite{meppelink2025hybrid}. 

Next, the proximity filter is applied to determine the proximity node set $\mathcal{V}^p$. For each $i \in \mathcal{V}^p$, \eqref{eq:sub_i} is solved to evaluate the effectiveness of splitting substation $i$ for congestion management. Since \eqref{eq:sub_i} only involves elements of substation $i$ and assumes other substations as single electric nodes, these problems can be solved very efficiently and in parallel for all $i \in \mathcal{V}^p$. The label is then assigned based on the prediction task.

\begin{algorithm}[t]
\caption{Training Data Generation}
\label{alg:data}

\For{data sample $s \in \Omega^s$}{

Sample inputs: $\hat{P}_{d,s},\hat{Q}_{d,s}$ $d\in \mathcal{D}$, outages $l \in \Omega^o_s$, generation costs $c_g$ $g \in \mathcal{G}_s$\;

Solve LP-AC-OPF to obtain $\hat{P}_g,\hat{Q}_g$ $g\in \mathcal{G}$\;

Compute base congestion cost $O^{CM}_0$\;

\If{no congestion}{
continue to next sample\;
}

Apply proximity filter to obtain $\mathcal{V}^p$\;

\For{$i \in \mathcal{V}^p$}{
Solve \eqref{eq:sub_i} for $z_i, z_{e\in E_i}, \mathcal{O}^{CM}_i$\;

Compute congestion reduction $\Delta \mathcal{O}^{CM}_i$\;

Assign label $y_i$ according to \cref{sec:task}\;
}
}
\end{algorithm}

\section{Case Studies}
\label{sec4:results}

\subsection{Setting and Test Networks}
The case studies are performed on small to large-scale IEEE and GOC test networks from Power Grid Library \cite{IEEEnetwork2019}, assuming the substation model in \cref{fig:substation}. The thermal limits of the IEEE 118-bus network is reduced to 80\% to induce congestion. Demand points are sampled within ±20\% of the nominal load, following a Kumaraswamy(1.6, 2.8) distribution with a Pearson correlation coefficient of 0.75 to capture load correlations \cite{giraud2024constraint}. For topology variations, up to N–2 planned line outages are sampled uniformly from non-radial lines. Generation costs are also sampled uniformly within ±20\% from nominal values to reflect flow redistribution and potential generator outages. Initial dispatches and flows are determined by solving a cost minimizing OPF. The proximity filter assumes $k=5$ hops. $\gamma=2$ and $s_{th}=0.8$ are used in \eqref{eq:CO}. For each N–k case, 10K samples are generated under DC PF, and 10K samples are generated for the N-0 case under LP-AC PF. Non-congested and infeasible samples are excluded from the dataset. All optimizations are modeled in Pyomo and solved using Gurobi 10.0 with Python 3.12.

The dataset is randomly divided into $70\%$ training, $10\%$ validation, and $20\%$ testing sets. ML models are implemented using PyTorch 2.5.0 and PyG 2.6.1. For classifier training, binary cross-entropy loss with class weighting is used. The output threshold is set to 0.5, and $\mathcal{O}^{th}=0.05$ is used for labeling, 
though these thresholds can be adjusted. For regression training, the mean squared error (MSE) loss and $\Phi(\cdot)=\text{clip}(\Delta\mathcal{O}^{CM}_i, -0.2,\infty)$ are considered. All models are trained with the Adam optimizer up to 100 epochs at a learning rate of $10^{-3}$ with early stopping. Training and testing are performed on the DelftBlue supercomputer using 8 CPU cores (Intel Xeon E5-6248R 3.0 GHz), 32 GB RAM and NVIDIA V100S GPUs. For case studies, all models are run over 3 random seeds for data shuffling and weight initialization. The reported results are the average over these runs. 


The ML models considered include an \textbf{MLP}, which uses concatenated node and edge features $x_i,e_{i,j}$ and one-hot encoding for line outages, and several GNN-based models with 5 layers and 64-dimensional hidden size: \textbf{Hom-GCN} (homogeneous graph convolutions), \textbf{Hom-MPNN} (homogeneous message passing \cref{sec:MPNN}), and \textbf{Het-MPNN} (heterogeneous message passing \cref{sec:HetGNN}). All GNNs include the encoders and the splitting decoder in \cref{fig:architecture}.

The following NTO-CM approaches are compared:
\begin{itemize}
\item \textbf{Org-NTO}: the original NTO-CM optimization in \eqref{eq:CO} which serves as the ground-truth.
\item \textbf{5-hop-NTO}: busbar splitting limited to substations within $k=5$ hops of congested lines.
\item \textbf{No-SW}: no busbar splitting (switching) is applied.
\item \textbf{Clf-GNN-NTO}: the proposed GNN-NTO using the top-5 actions predicted by the classification Het-MPNN.
\item \textbf{Reg-GNN-NTO}: the proposed GNN-NTO using the top-5 actions predicted by the regression Het-MPNN.
\end{itemize}

\subsection{Prediction Accuracy}

\cref{table:acc} compares the performance of different ML models in predicting the busbar splitting status of substations. Among the Clf evaluation metrics, accuracy can be misleading due to class imbalance (a model predicting mostly ‘no-split’ may still achieve high accuracy). F1-score balances precision and recall, reflecting how effectively the model identifies true splitting actions. The GCN shows the weakest performance due to its neglect of edge features. The MPNNs outperform the MLP, particularly in the larger GOC 2000-bus system by achieving an F1-score of 0.95 and an accuracy of 0.98. In other words, the fully-local MPNN with only 5 layers of message passing outperforms the \textit{global} MLP with more parameters, which confirms the locality of the NTO-CM problem and shows that MPNNs scale more efficiently while improving performance by learning local PF patterns. Finally, the Hom-MPNN and Het-MPNN perform similarly, as effective splitting actions (on substation level) depend primarily on local flow patterns rather than inter-type relations.


\begin{table}[t]
\renewcommand{\arraystretch}{1.0}
\centering
\caption{Performance comparison for N-0 cases.}
\label{table:acc}
\begin{tabular}{@{}clcccc@{}}
\toprule
                          &          & F1 Score & Accuracy & Precision & Recall \\ \midrule
\multirow{4}{*}{ieee-118} & MLP      & 0.87     & 0.93     & 0.81      & 0.94   \\
                          & Hom-GCN  & 0.84     & 0.91     & 0.77      & 0.92   \\
                          & Hom-MPNN & 0.90     & 0.95     & 0.85      & 0.97   \\
                          & Het-MPNN & 0.90     & 0.95     & 0.85      & 0.95   \\ \midrule
\multirow{4}{*}{ieee-300} & MLP      & 0.87     & 0.91     & 0.83      & 0.91   \\
                          & Hom-GCN  & 0.78     & 0.85     & 0.72      & 0.86   \\
                          & Hom-MPNN & 0.91     & 0.94     & 0.88      & 0.94   \\
                          & Het-MPNN & 0.91     & 0.94     & 0.86      & 0.95   \\ \midrule
\multirow{4}{*}{goc-500}  & MLP      & 0.85     & 0.94     & 0.77      & 0.96   \\
                          & Hom-GCN  & 0.79     & 0.91     & 0.68      & 0.95   \\
                          & Hom-MPNN & 0.89     & 0.96     & 0.83      & 0.97   \\
                          & Het-MPNN & 0.90     & 0.96     & 0.82      & 0.98   \\ \midrule
\multirow{4}{*}{goc-2000} & MLP      & 0.88     & 0.94     & 0.82      & 0.95   \\
                          & Hom-GCN  & 0.86     & 0.94     & 0.77      & 0.96   \\
                          & Hom-MPNN & 0.95     & 0.98     & 0.91      & 0.98   \\
                          & Het-MPNN & 0.95     & 0.95     & 0.92      & 0.98   \\ \bottomrule
\end{tabular}

\end{table}

\subsection{Generalization to Topology Changes}

\cref{table:top} reports F1 scores and compares model generalization under topology changes. We consider two scenarios: (I) zero-shot generalization to unseen topologies, by training and testing on different N-k cases (e.g., training on N-0 samples and testing on N-1 samples), and (II) training on a combined dataset of N-0, N-1, and N-2 samples. For the IEEE-14 system, all models experience a performance drop from N-0 to N-1, while the drop from N-1 to N-2 is smaller for MPNN models. This is because MPNNs can capture topological features from N-1 samples, improving generalization to N-2 cases. As system size increases, the zero-shot performance drop diminishes, as line outages in larger grids have a smaller relative impact, and models experience more diverse local topologies during training. As expected, performance on the combined dataset is higher than in the zero-shot cases. The proposed MPNN enhances model expressiveness over baseline models, which improves generalization to topology changes and superior performance once trained on different topologies. This benefit is particularly pronounced in large-scale systems where topology changes frequently occur in practice. 


\begin{table}[t]
\renewcommand{\arraystretch}{1.0}
\centering
\caption{Generalization to N-k topology changes.}
\label{table:top}
\begin{tabular}{clccccc}
\toprule
                          & Train set & N-0  & N-0  & N-1  & N-1  & N-0, N-1, N-2 \\
                          & Test set  & N-0  & N-1  & N-1  & N-2  & N-0, N-1, N-2 \\ \hline
\multirow{4}{*}{ieee-14}  & MLP       & 0.97 & 0.79 & 0.93 & 0.79 & 0.92        \\
                          & Hom-GCN   & 0.97 & 0.70 & 0.92 & 0.77 & 0.93        \\
                          & Hom-MPNN  & 0.98 & 0.80 & 0.95 & 0.85 & 0.96        \\
                          & Het-MPNN  & 0.98 & 0.81 & 0.96 & 0.86 & 0.96        \\ \midrule
\multirow{4}{*}{ieee-118} & MLP       & 0.87 & 0.79 & 0.82 & 0.79 & 0.83        \\
                          & Hom-GCN   & 0.84 & 0.80 & 0.80 & 0.78 & 0.82        \\
                          & Hom-MPNN  & 0.90 & 0.86 & 0.87 & 0.86 & 0.89        \\
                          & Het-MPNN  & 0.91 & 0.88 & 0.88 & 0.87 & 0.90        \\ \midrule
\multirow{4}{*}{ieee-300} & MLP       & 0.87 & 0.85 & 0.85 & 0.84 & 0.87        \\
                          & Hom-GCN   & 0.78 & 0.78 & 0.78 & 0.77 & 0.80        \\
                          & Hom-MPNN  & 0.91 & 0.90 & 0.89 & 0.88 & 0.90        \\
                          & Het-MPNN  & 0.91 & 0.90 & 0.89 & 0.88 & 0.91     \\ \bottomrule
\end{tabular}
\end{table}

\subsection{Transferability across Systems}

\cref{table:sys-gen} investigates the generalization and transferability of models across different power systems. We consider three cases: (I) \textbf{zero-shot (ZS)}: the GNN is trained on one system and tested on a different system without any retraining; (II) \textbf{transfer learning (TL)}: the GNN is first trained on the source system, and then only the encoders and decoder are retrained using 10\% of the samples from the target system (this setup reflects a realistic scenario where generating training data for large systems is time-consuming); (III) \textbf{combined dataset (CD)}, the GNN is trained on a merged dataset from both source and target systems and evaluated on the target system. Training and testing samples include all N-0, N-1, and N-2 samples for each system. The MLP is excluded here due to different input and output dimensionality across systems. 

Each power system has a distinct input and output distribution, posing a fundamental challenge for ZS transferability, as models trained on one system fail to generalize directly to another. However, in the TL case, retraining only the encoder and decoder with a limited number of samples effectively mitigates the distribution shift, while preserving the learned GNN embeddings. As system size increases, models encounter more diverse topologies and feature patterns, which improves TL performance, achieving an F1-score of up to 0.92 for the GOC-500 and GOC-2000 systems. Finally, the proposed MPNN achieves the best results in the CD case, highlighting its capacity to capture local PF patterns that improve cross-system generalization. This result shows that, instead of generating extensive training data for large-scale grids (which is computationally expensive), one can efficiently train a universal GNN using data from decomposed, congestion-prone regions of the grid.

\begin{table}[t]
\renewcommand{\arraystretch}{1.0}
\centering
\caption{Transferability across systems}
\label{table:sys-gen}
\begin{tabular}{@{}cclccc@{}}
\toprule
\begin{tabular}[c]{@{}c@{}}Source\\      System\end{tabular} & \begin{tabular}[c]{@{}c@{}}Target\\      System\end{tabular} &  & \begin{tabular}[c]{@{}c@{}}Zero-\\      Shot\end{tabular} & \begin{tabular}[c]{@{}c@{}}Transfer\\      Learning\end{tabular} & \begin{tabular}[c]{@{}c@{}}Combined\\      Dataset\end{tabular} \\ \midrule
\multirow{4}{*}{ieee-14} & \multirow{4}{*}{ieee-118} & MLP & - & - & - \\
 &  & Hom-GCN & 0.39 & 0.78 & 0.85 \\
 &  & Hom-MPNN & 0.28 & 0.86 & 0.90 \\
 &  & Het-MPNN & 0.30 & 0.86 & 0.90 \\ \midrule
\multirow{4}{*}{ieee-118} & \multirow{4}{*}{ieee-300} & MLP &  &  &  \\
 &  & Hom-GCN & 0.28 & 0.77 & 0.80 \\
 &  & Hom-MPNN & 0.39 & 0.87 & 0.90 \\
 &  & Het-MPNN & 0.31 & 0.87 & 0.91 \\ \midrule
\multirow{4}{*}{goc-500} & \multirow{4}{*}{goc-2000} & MLP &  &  &  \\
 &  & Hom-GCN & 0.32 & 0.83 & 0.85 \\
 &  & Hom-MPNN & 0.30 & 0.91 & 0.93 \\
 &  & Het-MPNN & 0.28 & 0.92 & 0.95 \\ \bottomrule 
\end{tabular}
\end{table}

\subsection{Computational Efficiency}
\cref{table:time} compares the computational efficiency of different NTO-CM approaches over three cases: DC PF with maximum 1 splitting, DC PF with maximum 3 splittings, and AC PF with maximum 1 splitting. The MIPGap in Gurobi is set to 1\%, and the time limit is 10 hours. Top 5 predictions are used as candidates in GNN-NTO. The high optimality gap of the No-SW indicates the effectiveness of busbar splitting in reducing grid congestion (up to 10\%). The negligible optimality gap of under 1\% for the 5-hop-NTO confirms that considering the proximity filter and exploiting the locality property of the NTO-CM problem is effective, especially in larger systems where solving the full optimization becomes intractable. The GNN-NTO approaches achieve substantial computational speed-ups, up to 4 orders of magnitude for the GOC-2000 system. In this context, the Reg model achieves slightly lower optimality gaps than the Clf model, as it can better prioritize impactful splitting actions using the proposed congestion reduction index. However, this improved ranking performance comes with increased sensitivity to hyperparameters and training design. The practical differences between the Clf and Reg tasks are discussed in the next section. In the multiple switching case, the GNN-NTO optimality gap increases up to maximum 6\% as the GNN is trained to predict the effectiveness of a single splitting action and may miss multiple cooperative actions. Finally, the computational advantage of GNN-NTO becomes even more pronounced for large-scale systems under AC PF, where even the proximity-filtered optimization is intractable; in such cases, GNN-NTO can still provide feasible solutions within a minute and 2.3\% optimality gap. With more powerful hardware resources and efficient implementation (e.g., in lower-level languages or on GPUs), MIP process can be further reduced to a few seconds, achieving near-real-time performance.

\begin{table*}[t]
\renewcommand{\arraystretch}{1.0}
\centering
\caption{Computational efficiency of different NTO-CM approaches.}
\label{table:time}
\begin{tabular}{cl|ccc|ccc|ccc}
\toprule
  &  & \multicolumn{3}{c|}{DC 1-split} & \multicolumn{3}{c|}{DC 3-split} & \multicolumn{3}{c}{AC 1-split} \\
 &  & Time & Speed up (×) & \multicolumn{1}{c|}{Gap (\%)} & Time & Speed up (×) & \multicolumn{1}{c|}{Gap (\%)} & Time & speed up (×) & gap (\%) \\  \midrule
\multirow{5}{*}{ieee-118} & Org-NTO & 18.3 s & - & - & 2.2 min & 1× & 0.0 & 1.5 min & - & - \\
 & 5-hop-NTO & 18.3 s & 1× & 0.4 & 45.7 s & 3× & 0.8 & 9.5 s & 10× & 0.5 \\
 & No-SW & 0.4 s & 46× & 8.9 & 0.5 s & 283× & 23.1 & 0.6 s & 149× & 6.7 \\
 & Clf-GNN-NTO & 0.9 s & 21× & 2.2 & 5.2 s & 26× & 7.3 & 2.7 s & 34× & 3.3 \\
 & Reg-GNN-NTO & 1.3 s & 14× & 1.7 & 10.0 s & 13× & 2.2 & 3.2 s & 29× & 1.2 \\ \midrule
\multirow{5}{*}{ieee-300} & Org-NTO & 49.8 s & - & - & 12.1 min & 1× & 0.0 & 3.4 min & - & - \\
 & 5-hop-NTO & 24.0 s & 2× & 0.4 & 8.1 min & 1× & 0.7 & 2.0 min & 2× & 0.2 \\
 & No-SW & 0.6 s & 84× & 10.2 & 0.8 s & 949× & 41.1 & 1.5 s & 140× & 1.4 \\
 & Clf-GNN-NTO & 1.5 s & 33× & 5.5 & 6.6 s & 110× & 12.1 & 4.0 s & 51× & 0.4 \\
 & Reg-GNN-NTO & 2.1 s & 24× & 1.1 & 22.9 s & 32× & 6.0 & 9.3 s & 22× & 0.2 \\ \midrule
\multirow{5}{*}{goc-500} & Org-NTO & 4.1 min & - & - & 35.6 min & 1× & 0.0 & 46.7 min & - & - \\
 & 5-hop-NTO & 2.1 min & 2× & 0.8 & 5.9 min & 6× & 3.2 & 17.2 min & 3× & 0.0 \\
 & No-SW & 1.1 s & 230× & 2.5 & 1.2 s & 1838× & 10.6 & 3.9 s & 712× & 1.7 \\
 & Clf-GNN-NTO & 2.3 s & 108× & 1.9 & 11.0 s & 194× & 5.1 & 8.9 s & 314× & 2.6 \\
 & Reg-GNN-NTO & 4.1 s & 61× & 0.7 & 17.8 s & 120× & 2.5 & 19.1 s & 147× & 2.1 \\ \midrule
\multirow{5}{*}{goc-2000} & Org-NTO & $>$10.0 hr & - & - & $>$10.0 hr & - & - & $>$10.0 hr & - & - \\
 & 5-hop-NTO & 3.4 hr & $>$3× & -9.2 & $>$10.0 hr & $>$1× & -3.0 & $>$10.0 hr & $>$1× & -6.7 \\
 & No-SW & 5.5 s & $>$6510× & 0.0 & 6.3 s & $>$5735× & 0.0 & 14.0 s & $>$2573× & 0.0 \\
 & Clf-GNN-NTO & 20.0 s & $>$1799× & -4.3 & 38.3 s & $>$940× & -0.5 & 54.0 s & $>$666× & -2.8 \\
 & Reg-GNN-NTO & 11.8 s & $>$3049× & -6.3 & 2.5 min & $>$244× & -1.1 & 1.3 min & $>$462× & -4.4 \\ \bottomrule
\end{tabular}
\end{table*}

\subsection{Discussion}
The presented case studies demonstrate promising results for the proposed GNN-NTO approach for congestion management. \cref{table:acc} shows that the MPNNs improve prediction accuracy compared to other architectures, while the heterogeneous GNN model does not provide additional benefit for the node candidate prediction task in this work. \cref{table:top} shows that generalization to topology changes improves in larger systems, although zero-shot generalization to unseen topologies remains limited. \cref{table:sys-gen} shows that the local formulation of the learning task together with the MPNN design improves transferability across systems with only minimal additional training or by training a universal model, while zero-shot generalization to entirely unseen systems remains a challenge. Finally, \cref{table:time} demonstrates the substantial computational speed-up achieved by the GNN-NTO approach while maintaining relatively small optimality gaps. However, some limitations of our approach can be noted.

Regarding the learning tasks, the Reg provides a more informative signal for ranking candidate substations. However, it is also more sensitive to the choice of hyperparameters $\gamma$ and $\Phi$ that define the congestion reduction index. Large values of $\gamma$ emphasize highly congested lines but may increase sensitivity to small variations in the grid state. In addition, the regression index can take negative values when a split worsens congestion, which can bias the learning objective if not properly controlled. Therefore, clipping in $\Phi$ is used for stable training. In this work, the regression formulation is explored as a proof-of-concept for predicting congestion reduction magnitudes to improve candidate selection. Further work is needed to refine this approach.

The GNN predicts the effectiveness of a busbar splitting action while assuming that no other substations are split. Consequently, when multiple splitting actions are required to mitigate congestion, the current approach identifies them sequentially, which does not guarantee finding the optimal combination. In addition, although the GNN-NTO approach provides significant computational speed-ups during online use, generating labeled training data remains computationally expensive for large-scale systems. Improving zero-shot transferability can address this limitation by enabling models trained on smaller systems to be applied directly to larger systems.

\section{Conclusion and Future Work}
\label{sec5:conc}
This paper presents a GNN-accelerated approach (GNN-NTO) for congestion management via busbar splitting. The proposed approach significantly reduces computations, enabling near-real-time solutions with operational feasibility. To this end, we develop a linearized AC NTO formulation to reduce grid congestion. A novel edge-aware MPNN is proposed to predict effective nodes for busbar splitting actions, which serve as candidate solutions for the MIP problem.
A proximity filter further improves learning performance by focusing the prediction task on the most relevant local regions. Two learning tasks are explored: a classification task that predicts whether splitting a substation alleviates congestion, and a regression task that estimates the congestion reduction associated with each splitting action. While the regression formulation can improve the prioritization of candidate nodes and provide interpretability, it is also more sensitive to hyperparameter choices and requires further investigation. Case studies demonstrate that the proposed MPNN learns generalizable features from local PF patterns. In particular, the results show that generalization to topology changes improves in larger systems, and that transfer learning across systems is effective with only minimal additional training, allowing a pre-trained GNN to be efficiently adapted to new grids or enabling a universal model to be trained across multiple systems. These properties are particularly valuable for large-scale systems with frequent topology changes, varying grid conditions, and expensive data collection. However, zero-shot generalization to entirely unseen systems remains challenging due to distributional shifts in the input and output spaces. Moreover, the GNN-NTO achieves up to 4 orders of magnitude speed-ups for the large-scale GOC 2000-bus system under AC PF, providing a feasible solution in one minute with a minor optimality gap. 

Future research will focus on zero-shot generalization to unseen grids, addressing distributional shifts in ML-based optimization proxies, and extending the approach for multiple busbar splitting actions.

\section*{Acknowledgment}
This work is supported by the TU Delft AI Labs \& Talent Programme. The authors are thankful to Olayiwola Arowolo from Delft University of Technology who provided valuable discussions.

\newcommand{\BIBdecl}{\setlength{\itemsep}{-0em}}

\bibliographystyle{IEEEtran}
\bibliography{Library}

\end{document}